\documentclass[10pt, a4paper]{article}

\usepackage[]{lrec-coling2024} % this is the new style
\usepackage[nonumberlist, nopostdot, hyperfirst=false]{glossaries}
\glsdisablehyper
\usepackage{booktabs}
\usepackage{framed}

\def\exwrds{\emph{extreme-words}}
\def\strwrds{\emph{strong-words}}

\newacronym{ica}{ICA}{Independent Component Analysis}
\newacronym{nlp}{NLP}{Natural Language Processing}
\newacronym{nmt}{NMT}{Neural Machine Translation}
\newacronym{pca}{PCA}{Principal Component Analysis}
\newacronym{pos}{POS}{Part of Speech}

\title{\textbf{Exploring Interpretability of Independent Components of Word Embeddings with Automated Word Intruder Test}}

\name{Tomáš Musil, David Mareček} 

\address{
         Charles University, Faculty of Mathematics and Physics,\\
         Institute of Formal and Applied Linguistics \\
         \{musil,marecek\}@ufal.mff.cuni.cz\\}

\abstract{
Independent Component Analysis (ICA) is an algorithm originally developed for finding separate sources in a mixed signal, such as a recording of multiple people in the same room speaking at the same time. Unlike Principal Component Analysis (PCA), ICA permits the representation of a word as an unstructured set of features, without any particular feature being deemed more significant than the others. In this paper, we used ICA to analyze word embeddings. We have found that ICA can be used to find semantic features of the words, and these features can easily be combined to search for words that satisfy the combination. We show that most of the independent components represent such features. To quantify the interpretability of the components, we use the word intruder test, performed both by humans and by large language models. We propose to use  the automated version of the word intruder test as a fast and inexpensive way of quantifying vector interpretability without the need for human effort.
 \\ \newline  }

\begin{document}

\maketitleabstract

\section{Introduction}

This paper centers on the exploration of word embeddings through the lens of \gls{ica}. Unlike \gls{pca}, \gls{ica} permits the representation of a word as an unstructured set of features, without any particular feature being deemed more significant than the others. Essentially, we view the vector representations of words as a combination of interpretable features, and our goal is to identify these features.

The main contribution of the paper is interpretability. Because ICA is a linear transformation of the embedding vectors, we do not expect any change in the results of downstream tasks. Although more interpretable representations will not help the model performance, they may help us understand how the tasks are performed by the models and what information is stored in the embeddings. In addition to theoretical implications, this also impacts trust in the models used in practice.

We show that most of the \gls{ica} components can be interpreted and the interpretable components can be combined to find words that have the features associated with both components. To quantify the interpretability, we use the word intruder test, both with humans and with large language models.

\begin{figure*}[t]
    \centering

\includegraphics[width=0.33\textwidth]{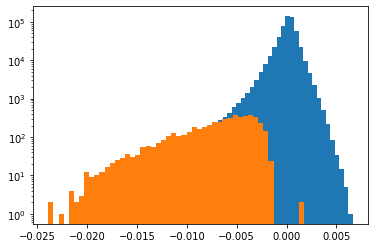}\includegraphics[width=0.33\textwidth]{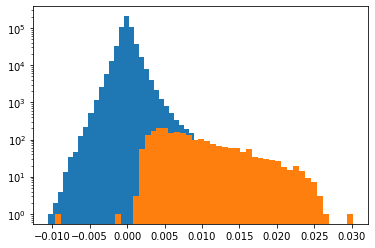}\includegraphics[width=0.33\textwidth]{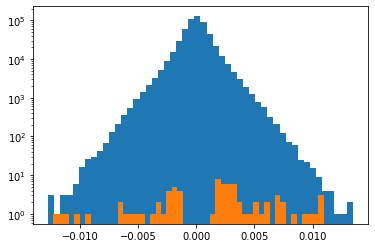}

    \caption{Histograms of distributions of words along a particular component.
	Orange bars represent \strwrds{}. Blue bars represent
	the rest of the vocabulary. Note that the vertical axis is logarhitmic,
	otherwise the orange bars would be too low to be distinguishable. There
	are three typical shapes of these histograms: orange mass in the
	negative direction, orange mass in the positive direction, and 
	small amount of orange scattered randomly. This shows that the components 
	usually capture some feature in one direction, which is arbitrary (property
	of the ICA algorithm), or contain random noise.}

    \label{fig:stran}
\end{figure*}

\section{Independent Component Analysis}
\label{sec:ica}

\Gls{ica} \cite{comon_independent_1994} is an algorithm originally developed for finding separate sources in a mixed signal, such as a recording of multiple people in the same room speaking at the same time. In the past, it was also used for automatic extraction of features of words \cite{honkela_wordicaemergence_2010}. 

The \gls{ica} algorithm \citep{hyvarinen_independent_2000} consists of:
\begin{enumerate}
    \item optional dimension reduction, usually with \gls{pca},
    \item centering the data (setting the mean to zero) and \emph{whitening} them (setting variance of each component to 1),
    \item iteratively finding directions in the data that are the most non-Gaussian.
\end{enumerate}

The last step is based on the assumption of the central limit theorem: the
mixed signal is a sum of independent variables, therefore it should be closer
to the normal distribution than the variables themselves.

The \gls{ica} algorithm is stochastic; every run gives a slightly different result. It always returns as many components as we specify before running it (up to the dimension of the original data). If the data was generated by a lower number of independent components and some random noise, \gls{ica} will return some components containing only the noise.

\gls{ica} may be an interesting tool for analysis of word embeddings also from a theoretical point of view.
Following \citet{musil-2021-representations}, we believe that it might be useful to
conceptualize meaning of an expression as a combination of various components.
These components emerge from the use of the expression in context.
Each of them would represent a specific relation to other expressions, forming a continuous structure that does not adhere to a simple tree hierarchy.

ICA of word embeddings is a plausible candidate for such conceptualization because it allows us to represent a word as an unstructured set of features, without some of them being necessarily more important than others.

In contrast, the commonly used \gls{pca} components are ordered by the amount of variance in the data explained by each component. This ordering can also be interpreted as a hierarchy, with, e.g., verbs versus nouns being a typical first component \citep{musil_examining_2019}, the following components separating adjectives and adverbs, later components separating modal verbs from the rest of the verbs, proper nouns from the rest of the nouns, etc.

\section{Experiments and Results}
\label{sec:exp}

Most of our experiments were carried out on the One Billion Word Benchmark corpus
\citep{chelba2013one}. The corpus mostly contains data from the news and parliamentary proceedings domains.
To show examples from a different text domain, we have also used the English side of the section \emph{c-fiction} of the CzEng~1.7 corpus \cite{bojar_czeng_2016}, containing 78M tokens (997k unique tokens) of short passages from various fiction books.

We have trained word2vec \cite{mikolov_efficient_2013} embeddings on the corpus
with 512 dimensions (skip-gram with negative sampling, window size 10), and ran the \gls{pca} and \gls{ica}\footnote{We are using the scikit-learn \cite{pedregosa_scikit-learn_2011} implementation
of the FastICA algorithm \cite{hyvarinen_fast_1999}.} (into 512 components) on them.

Due to the random initialization, each run of \gls{ica} produces a slightly different result.
To assess the consistency of \gls{ica}, we compared two independent runs of \gls{ica} performed on the same embeddings. For a large proportion of the components, a component from one run is strongly correlated to exactly one component from the other run.

To examine what each component represents, we can look at the words in the vocabulary that are the highest or the lowest in that particular component (we will call these the \exwrds{}). For a vocabulary $V$, where each word is associated with a $d$-dimensional vector representation $r: V \to R^d$ and $r(w)_c$ denoting the $c^{\text{th}}$ component of the representation of the word $w$, we can define the sets of $k$ \exwrds{} in positive and negative directions as:

$$E(c, +, k) = \{w \in V: |\{x \in V: r(x)_c \geq r(w)_c\}| \leq k\},$$
$$E(c, -, k) = \{w \in V: |\{x \in V: r(x)_c \leq r(w)_c\}| \leq k\}.$$

From a different point of view, we can also look at each word and determine which component is the \emph{strongest} (is the largest in absolute value) for that word and whether it is positive or negative. Thus, we are able to associate a particular component and direction with each word. And for each direction of each component, we can find a set of words for which this component/direction is the strongest one (see also \citet{honkela_wordicaemergence_2010}). We will call these \strwrds{}. Using the same notation as in the previous equations, we can define function $SC: V \to ([1,d], \{'+', '-'\})$ that assigns each word a \emph{strong} component and direction as:
$$SC(w) = (a := argmax_c(|r(w)_c|),$$
$$\text{'+' if }r(w)_a \geq 0 \text{ otherwise '-'})$$
and then define the set of \strwrds{} $S(c, dir)$ for component $c$ and positive/negative direction as:
$$S(c, dir) = \{w \in V:  SC(w) = (c, dir)\}.$$

\subsection{Component Directionality}
\label{sec:direction}

The distribution of words along a component usually follows a pattern: most
words are located around 0 and a smaller group of words is separated in either
the positive or the negative direction.
Figure~\ref{fig:stran} illustrates this pattern of uni-directionality by plotting the distribution of \strwrds{} along a component.
We see two characteristic patterns of this distribution:
either most of \strwrds{} are located in one of the positive/negative half-spaces, or there are relatively few \strwrds{} that are evenly distributed across both sides.

Our hypothesis is that the components that are one-sided are the ones that are interpretable, while the spread-out components mostly contain noise.

In our experiments on the components of the embeddings trained on the Billion corpus, there were approximately 361 one-sided components versus 161 spread-out components (averaged over multiple ICA runs), based on the ratio of words for which the component is the largest component (as in Figure~1; we count components with more than 70\% of ‘strong-words’ as ‘one-sided’).

\subsection{Word Intruder Test}

\begin{table}[ht]
    \centering
    \begin{tabular}{lccc}
    \toprule
    \textbf{vect.} & \textbf{i. identified} & \textbf{agr. i.}&\textbf{agr. non-i.}\\
    \midrule
    \multicolumn{4}{c}{Random baseline}\\
     & 204.8 (20\%) & 8.1 (1\%)& \textbf{8.1} (1\%)\\
     
    \midrule
    \multicolumn{4}{c}{Human}\\
    w2v & 317.3 (31\%) &  90 (9\%)&120 (12\%)\\
    ICA & \textbf{425.6} (42\%) & \textbf{190} (19\%)&82 (8\%)\\
    
    \midrule
    \multicolumn{4}{c}{GPT-3.5}\\
    w2v & 291.5 ($\pm$12.1) & -- &  -- \\
    PCA & 273.7 ($\pm$11.7) & -- &  -- \\
    ICA & \textbf{467.1} ($\pm$ 6.4) & -- & -- \\
    
    \midrule
    \multicolumn{4}{c}{GPT-4}\\
    w2v & 273  & -- & -- \\
    ICA & \textbf{524} & -- & -- \\
    \bottomrule
    \end{tabular}
    \caption{Results of the word intruder test on the Billion corpus. Percentages indicate the proportion of all of the components/directions. Ranges in parentheses indicate standard deviation over 5 randomized test sets. For word2vec dimensions, the intruder word was on average identified less often than for the ICA components. The annotator agreement on the correct intruder word is higher for the ICA components, as is the ratio between the number of cases where the annotators agreed on the correct intruder word against the number of cases where they agreed on a word that was not the correct intruder. This indicates that the ICA components are more interpretable than the original word2vec dimensions. We assume that most of the components are one-sided; therefore, the maximum amount of interpretable components is around 50\% (we test both directions, but assume only one is interpretable). Because every question contains 5 possible answers, there is a 20\% chance of guessing the correct intruder at random. Therefore the range of the interpretability score in Table 1 is between $\sim$20\% and $\sim$50\%, making the difference of 11\% quite large.
}
    \label{tab:intruder1}
\end{table}

To estimate intepretability of the components, we performed the \emph{word intruder test} \citep{NIPS2009_f92586a2}, that has been widely used for this purpose \citep{Subramania2018}. This test presents the annotators with 5 words, 4 of which are the 4 \exwrds{} of the tested component and direction. The fifth word is an \emph{intruder}, selected randomly from the top 10\% of words from another random component and direction. If the component is interpretable, the \exwrds{} should form a coherent set and the annotators should be able to identify the intruder.

We had the intruder test data (based on the \emph{Billion} corpus, see Section~\ref{sec:exp}) for word2vec and \gls{ica} components annotated by three independent annotators. The results in Table~\ref{tab:intruder1} show that \gls{ica} components are more interpretable than the components of original word2vec embeddings. The intruder test also shows that the components are usually interpretable only in one direction.

To avoid the high cost of manual annotation, we performed further intruder tests with the GPT-3.5 language model \citep{brown2020language}. In this setting, we were able to randomize the selection of the coherent set and pick 4 words at random from the 20 \exwrds{} for each component and direction. We generated 5 randomized tests for each representation/component/direction. We used the prompt ``Which word does not fit the following group of words? <$w_1$>, <$w_2$>, <$w_3$>, <$w_4$>, <$w_5$>. Answer using just one word."
Initially we chose to put the test words in the prompt in random order. However, we have noticed that the language model is biased to select words at certain positions more often than others. We solved this by repeating each test 5 times with the test words positions rotating, in order for the intruder word to occur in all 5 positions. We consider the intruder word detected correctly if at least 3 of the 5 rotated tests are answered correctly.

The results of intruder test with GPT-3.5 (Table~\ref{tab:intruder1}) are consistent with the manual tests. While word2vec components tested above random baseline, the ICA components have a significantly higher score. The number of test instances where the intruder was correctly identified by GPT3.5 correlates with the percentage of vocabulary that are \strwrds{} for the tested component/direction (Pearson's $r = 0.65$). This is consistent with our hypothesis that uni-directional components are interpretable.  We have also tested PCA components in this setting. The score for PCA was slightly lower than for word2vec. We think this is because PCA is constructed to fit the highest amount of information into the lowest possible number of components, leaving most of the components as random uninterpretable noise.

We also tested a limited number of examples with GPT-4 \citep{openai2023gpt4}. This larger model achieved significantly higher score on the ICA components intruder test, while the score for the word2vec components was similar to GPT-3.5.

\subsection{Combining the Components}
\def\comp#1{\textbf{#1}}
 
We can combine a pair of components by searching for words for which the product of the components is the highest.\footnote{As we have seen in Section~\ref{sec:direction}, each component is either positive, negative, or noisy. We can compute the mean value of \strwrds{} for each component and then flip the sign of that component if the mean is negative. In the rest of this section, we assume that this operation was carried out on the model and all of the components that represent semantic features do so in the positive direction.} For example, in our particular instance of ICA of word2vec embeddings of the English side of the CzEng-fiction corpus, the 15 words for which the value of  $C_{398}$ (component number 398) is the highest are the following:
\emph{rumble, booming, roar, wail, sound, murmur, shouts, cries, louder, shrill, screams, noises, muffled, voices, howl}. We see that this components has high values for words associated with \comp{sound}. For $C_{110}$, the top 15 words are associated with \comp{animals}: \emph{cats, predators, rats, predator, lions, fox, rabbits, bears, wolves, lion, deer, dogs, mice, tigers, cat}. If we search for the top 15 words for which $C_{398} \cdot C_{110}$ is the highest, we get the following:

\comp{sound} $\ast$ \comp{animals}:
\emph{growl, barking, purr, growls, whine, baying, growling, howl, yelp, bleating, chirping, buzzing, squealing, squeals, crickets}

Here are a few hand-picked examples from the same model:

\comp{sound} $\ast$ \comp{horses}:
\emph{hooves, hoofs, hoofbeats, snort, hoof, whinny, jingling, snorting, clop, clink, whinnying, thudding, jingle, shod, neighing}

\comp{sound} $\ast$ \comp{play}:
\emph{melody, flute, music, musical, chords, orchestra, guitar, stringed, violin, trumpets, tune, accompaniment, piano, Bach, melodies}

\comp{sound} $\ast$ \comp{door}:
\emph{click, clang, creak, thud, clanged, clank, clink, splintering, clunk, squeak, groan, audible, snick, thunk, footsteps}

\comp{clothing} $\ast$ \comp{army}:
\emph{fatigues, uniforms, regimental, insignia, Infantry, uniform, tabs, breastplate, vests, stripes, Kevlar, Armored, helmets, outfit, pants}

\comp{units} $\ast$ \comp{money}:
\emph{dollars, cent, cents, francs, bucks, per, dollar, billion, roubles, shillings, million, percent, guineas, pounds, pence}

We have also succesfully tested this with pairs of sports and countries on the Billion corpus.

\section{Discussion and Future Work} 

ICA can provide components that are interpretable without relying on predetermined set of categories. The resulting components may represent categories that are not very general and are perhaps not suitable as a general semantic representations to use in practical applications. They do not seem to represent semantic primitives as defined by \citet{article}. Examples from the ICA of word2vec embeddings trained on the Billion corpus, interpreted by looking at the \exwrds{} and finding what they have in common, include components that represent various sports, states, types of numbers (e.g. years, basketball scores, percentages; each have their own component) or a component representing surnames of famous people who's first name is David. This may not be very useful in general, but because the ICA can easily be mapped to the original embeddings, it shows how the information is organised in the embeddings and consequently in the corpus itself (in this case, large portion of the corpus consists of news articles). Furthemore, there is the possibility of combining the individual components.

Regarding lexical semantics, this work is connected with theories that use the notion of a ‘semantic feature’ and shows that we can empirically find this kind of structure in the embeddings. Our work presents a possible way to fix one of the shortcomings of componential analysis, that “The discovery procedures for semantic features are not clearly objectifiable”\footnote{https://en.wikipedia.org/wiki/Componential\_analysis}.
W.r.t the structure of the lexicon, this tells us that we can organize the lexicon by binary semantic features that words either have or do not have.

In future work, we are going to concentrate on automatically detecting the interpretation for each component and finding which components can be combined together, aiming at unsupervised construction of a compositional semantic map of word embeddings (and by extention also of the underlying text corpus). We believe that this may be useful not only for interpreting various forms of vector representations, but also as a method of computational analysis of compositional structures present in various corpora, as a form of ``distant reading" \citep{moretti2000conjectures}.

\gls{ica} could also be useful to identify potential for various biases in the representations (e.g. gender bias; see Appendix~\ref{app:exa} for examples). If there are components clearly showing structure related to a sensitive attribute associated with a word (such as gender role), there is a potential of misusing this information in a machine learning system that uses or generates the representations.
 
Based on experiments presented in this paper, it seems that the automated word intruder test with large language models is a viable alternative to other methods to quantify the interpretability of word vector representations without requiring human effort, such as the one proposed by \citet{SenelUYKC18}. The benefits of automated intruder test are its simplicity and possibility of directly comparing the results to human evaluation of the same test examples in cases where the human labour is available. More work needs to be done to determine under what conditions (specific prompts, language models, and other variables) is it possible for the automated word intruder test to be used reliably.

\section{Related Work}
\label{sec:rel}

\citet{vayrynen_comparison_2005} devised a method to quantify how well the
unsupervised features correspond to a set of linguistic features such as part of speech categories. They compared SVD and ICA on context-word matrix and concluded that ICA corresponds better to human intuition.

\citet{musil_examining_2019} examined the structure of word embeddings with
\gls{pca}. They found that \gls{pca} dimensions correlate strongly with
information about \gls{pos} and that the shape of the space is strongly
dependent on the task for which the network is trained.

\citet{faruqui-etal-2015-sparse} and \citet{Subramania2018} generated sparse interpretable representations from word embeddings. Unlike ICA, these are not simple projections of the original vectors.

Related work on the examination of vector representations in \gls{nlp} was surveyed by
\citet{bakarov_survey_2018}. More information can also be found in
the overview of methods for analysing deep learning models for \gls{nlp} by
\citet{belinkov_analysis_2019}. For more on interpretation in general
and unsupervised methods in examining word embeddings, see 
\citet%[Chapters 3 and 4]
{marecek_hidden_2020}.

\section{Conclusion}

\gls{ica} components correspond to various features, that seem to be mostly semantic. These features tend to be binary and the components are unidirectional.  We have demonstrated that components can be combined as semantic features by simple multiplication, giving high values to words that combine the semantic features associated with the components. To quantify the interpretability, we have successfully used the word intruder test with large language models.

\section*{Acknowledgements}
We have been supported by grant 23-06912S of the Czech Science Foundation. We have been using language resources and tools developed, stored, and distributed by the LINDAT/CLARIAH-CZ project of the Ministry of Education, Youth and Sports of the Czech Republic (project LM2018101).

\nocite{*}
\section{Bibliographical References}\label{sec:reference}

\bibliographystyle{lrec-coling2024-natbib}
\bibliography{paper}

\appendix

\section{Examples of Components}
\label{app:exa}

In this section, we present a few examples of components to show the type of information they represent. The description of each component contains the corpus on which the word2vec embeddings were trained, the id of the component (arbitrary number and direction) and whether the words presented are \emph{end-words} or \emph{strong-words} (see Section~\ref{sec:exp} of the paper).

\def\ukazka#1#2{%
\begin{leftbar}%
\noindent
\textbf{#1:}\\%
\emph{#2}%
\end{leftbar}%
}

The first three examples show components representing specific groups of words: names of people associated with ``David'', abbreviations for representatives in the US, popular music groups.

\ukazka{Billion C3- end-words}{David Copperfield Archuleta Beckham Nalbandian Plouffe Letterman Goliath Friehling Lammy Souter Blanchflower Vitter Duchovny Martinon McKiernan Adom Garrard Legwand Cronenberg Fincher Aardsma Buik Skrela Hasselhoff Petraeus Wyss Pogue Furnish Mamet}

\ukazka{Billion C4- end-words}{Reps. R-Mich D-Mich Rep. D-Pa R-Calif R-la D-Ohio D-Mo D-Calif Edolphus D-Md D-CA R-Pa D-Minn R-Texas D-Conn R-Maine D-Hawaii R-Va D-Ind D-Wis D-Wash D-N.D. R-Tex D-Ore R-Ga R-Iowa D-N.Y. D-N.J.}

\ukazka{Billion C10- end-words}{Tings Coldplay MGMT Rascal Kasabian Metallica Radiohead Linkin rockers Flatts Dizzee Raconteurs Interscope Nickelback Zeppelin Billboard Prodigy Billboard.com Leppard Verve Paramore Depeche album supergroup R.E.M. Beastie Weezer Gorillaz Glasvegas Stryder}

Component number 143 shows words representing people in the Billion corpus \citep{chelba2013one}. Notice that this component (and no other component in this particular set of embeddings) does not differentiating the words based on the associated gender roles. Compare this with component 73 from word2vec embeddings trained on English Wikipedia\footnote{Downloaded from \url{https://dumps.wikimedia.org/enwiki/20231020/}.}, which shows ordering according to gender (in the opposite directions). Understanding how is this type of information represented and under what conditions is it more prominent in the representations may help us prevent unwanted bias in the systems that use these representations as the first step of a machine learning pipeline.

\ukazka{Billion C143+ end-words}{motorcyclist cyclist hiker firefighter serviceman sailor soldier sufferer climber protester skier man worker airman scientist diver rider woman jogger businesswoman journalist diplomat shopper traveler teenager surfer pensioner attendee person holidaymaker}

\ukazka{Billion C143+ strong-words}{man woman journalist person teenager worker guy performer politician motorcyclist sailor resident pensioner businessman scientist banker musician kickboxer supporter shopper salesman coworker colleague staffer traveler athlete holidaymaker citizen diplomat reveller player}

\ukazka{Wikipedia C73+ end-words}{feminist headmistress Giveen abbess prioress lady suffragist Xaveria Petyarre nun feminists Pizan suffragette benefactress alumna Nardal chairwoman Țig Abbess actresses matron Bessola Sister Abrikosova regnant Alacoque Overstake Smeal needleworker Tyutcheva}

\ukazka{Wikipedia C73- end-words}{Jesse Harold Robert Ryusuke Arthur David Daniel Lukáš Woodie Guy Andreev Shintaro balding Bjorn Remo countryman Richard Stanfield Adam Frat Hamish Jason Seth Kelvin Michael Granollers Zorin Łukasz Hieronymus Kaspar}

\end{document}